\documentclass[a4paper,fleqn]{cas-dc}

\usepackage[authoryear,longnamesfirst]{natbib}

\usepackage{siunitx}
\usepackage{subcaption}
\usepackage{multirow}
\usepackage{booktabs}
\usepackage{xurl}
\usepackage{cleveref}

\begin{document}

\let\WriteBookmarks\relax
\def\floatpagepagefraction{1}
\def\textpagefraction{.001}

\shorttitle{Avalanche Mapping from SAR Imagery with Deep Learning}
\shortauthors{M. Gatti et al.}

\title [mode = title]{Large-Scale Avalanche Mapping from SAR Images with Deep Learning-based Change Detection}


\author[insubria1]{Mattia Gatti}[orcid=0009-0006-8314-7092]
\cormark[1]
\ead{mgatti3@uninsubria.it}
\credit{Conceptualization, Methodology, Supervision, Software, Investigation, Data curation, Visualization, Writing - original draft}

\author[insubria2]{Alberto Mariani}
\credit{Data curation, Validation, Formal analysis, Visualization, Writing - review \& editing}

\author[insubria1]{Ignazio Gallo}
\credit{Conceptualization, Methodology, Visualization, Writing - review \& editing, Supervision}

\author[alpsolut]{Fabiano Monti}
\credit{Resources, Data curation, Investigation, Writing - review \& editing}


\affiliation[insubria1]{
    organization={Department of Theoretical and Applied Sciences, University of Insubria},
    addressline={Via Ottorino Rossi 9},
    city={Varese},
    postcode={21100},
    country={Italy}
}

\affiliation[insubria2]{
    organization={Department of Science and High Technology, University of Insubria},
    addressline={Via Valleggio 11},
    city={Como},
    postcode={22100},
    country={Italy}
}

\affiliation[alpsolut]{
    organization={Alpsolut S.r.l.},
    addressline={Via Saroch 1098/A},
    city={Livigno},
    postcode={23041},
    country={Italy}
}


\cortext[1]{Corresponding author}

\begin{abstract}
Accurate change detection from satellite imagery is essential for monitoring rapid mass-movement hazards such as snow avalanches, which increasingly threaten human life, infrastructure, and ecosystems due to their rising frequency and intensity. This study presents a systematic investigation of large-scale avalanche mapping through bi-temporal change detection using Sentinel-1 synthetic aperture radar (SAR) imagery. Extensive experiments across multiple alpine ecoregions with manually validated avalanche inventories show that treating the task as a unimodal change detection problem, relying solely on pre- and post-event SAR images, achieves the most consistent performance. The proposed end-to-end pipeline achieves an F1-score of 0.8061 in a conservative (F1-optimized) configuration and attains an F2-score of 0.8414 with 80.36\% avalanche-polygon hit rate under a less conservative, recall-oriented (F2-optimized) tuning. These results highlight the trade-off between precision and completeness and demonstrate how threshold adjustment can improve the detection of smaller or marginal avalanches. The release of the annotated multi-region dataset establishes a reproducible benchmark for SAR-based avalanche mapping.
\end{abstract}


\maketitle


\section{Introduction}
\label{sec:intro}

Snow avalanches are fast-moving natural hazards that threaten life, infrastructure, and ecosystems in mountainous regions~\citep{eaws_fatalities}. Rapid and reliable mapping is therefore important for hazard assessment and post-event documentation, especially as climate change is expected to affect avalanche activity and related mass-movement hazards~\citep{Dubey2023MassMovement}. In practice, avalanche information is often derived from field observations and incorporated into regional bulletins issued by the European Avalanche Warning Services (EAWS)~\citep{eaws_standards}. However, field-based observations are limited by poor visibility, harsh weather, and the inaccessibility of steep terrain~\citep{nhess-2024-48}.

Sentinel-1 synthetic aperture radar (SAR) imagery is well suited to avalanche monitoring because it can be acquired independently of daylight and cloud cover~\citep{Kapper2023AvalancheMonitoring}. Fresh avalanche deposits and roughened tracks often appear as localized backscatter changes between pre- and post-event images, making the problem naturally amenable to bi-temporal change detection~\citep{asokan2019change,KESKINEN2022103558}. At the same time, avalanche mapping from SAR remains challenging because similar backscatter variations can also be produced by snowfall, wind redistribution, melt-refreeze processes, and other surface changes.

Many existing workflows address this difficulty by incorporating terrain-derived information or manually engineered rules. While useful, these additions increase preprocessing complexity and can limit scalability in large-area or near-operational settings. This raises a practical question: are auxiliary terrain inputs truly necessary, or can a modern change-detection model learn sufficiently informative representations directly from bi-temporal SAR imagery?

This study addresses that question through a systematic investigation of large-scale avalanche mapping with Sentinel-1 SAR. The central finding is that a unimodal formulation based only on pre- and post-event SAR images is sufficient and shows more consistent performance than multimodal alternatives that include terrain auxiliaries. Building on this result, an end-to-end pipeline is developed for large-scale inference and operationally relevant threshold selection.

The main contributions are threefold:
(i) a systematic comparison of multimodal and unimodal change-detection strategies for avalanche mapping, showing that the most reliable solution is a unimodal SAR-only formulation;
(ii) the release of a manually validated multi-region avalanche inventory geodatabase to support reproducible benchmarking; and
(iii) an analysis of patch blending and threshold selection strategies, including both F1- and F2-oriented tuning, to characterize the precision-recall trade-off during large-scale inference.

To support transparency and reproducibility, the annotated dataset and code are publicly available (see Code Availability and Data Availability sections). The next section reviews prior work in SAR-based avalanche detection and deep learning for remote sensing change detection.

\section{Related Work}
\label{sec:related}

Sentinel-1 SAR has become a key data source for avalanche monitoring due to its ability to acquire imagery independently of cloud cover and daylight conditions. Early SAR-based approaches for mass-movement mapping typically relied on manual interpretation or simple threshold-based segmentation of backscatter differences, which limited their scalability and robustness for large-area monitoring~\citep{stumpf2011object}. To improve automation, several studies introduced feature-engineering workflows combining backscatter change metrics, terrain masks, object-based segmentation, and classical machine-learning classifiers. For example, Waldeland et al.~\citep{8517536} combined SAR difference imaging with a convolutional classifier to refine avalanche detection, while Hamar et al.~\citep{7729173} proposed a two-stage approach in which candidate regions were first identified through backscatter differencing and subsequently classified using a random forest with both SAR-derived and DEM-based features.

More recently, deep learning has emerged as a powerful framework for change detection in remote sensing. Early convolutional architectures, such as Siamese networks~\citep{daudt2018fully} and STANet~\citep{Chen2020}, established the paradigm of jointly processing pre- and post-event imagery to learn change-relevant representations. Subsequent research introduced attention mechanisms and transformer-based architectures to better capture spatial and temporal dependencies. For instance, BIT~\citep{chen2021remote} and SNUNet-CD~\citep{fang2021snunet} incorporated token-based attention and nested U-Net structures for improved feature fusion, while transformer-based models such as ChangeFormer~\citep{bandara2022transformer} leveraged multi-scale self-attention to model long-range interactions in change detection tasks. Lightweight architectures have also been explored; TinyCD~\citep{codegoni2023tinycd} demonstrated that carefully pruned U-Net designs can achieve competitive performance with significantly fewer parameters. More recently, STNet~\citep{10219826} further improved boundary precision through refined cross-temporal and multi-scale feature fusion, and the Open-CD framework~\citep{li2024open} unified many of these architectures within a common experimental platform.

Most change detection architectures rely primarily on the appearance of optical or SAR imagery and do not explicitly incorporate auxiliary geophysical information. In classical avalanche mapping, however, terrain context and radiometric effects have traditionally played an important role in interpretation. Avalanche deposits typically occur under specific terrain conditions, such as particular slope and aspect ranges~\citep{sovilla2010variation}, while the local incidence angle (LIA) strongly influences the backscatter intensity observed in SAR imagery~\citep{KESKINEN2022103558}. For this reason, terrain-derived predictors and radiometric normalization are often considered essential components of operational workflows.

This assumption has motivated several recent studies exploring multimodal deep learning models that integrate terrain information alongside satellite imagery. For example, Lu et al.~\citep{lu2023dual} proposed a dual-encoder U-Net architecture for landslide detection that fuses Sentinel-2 imagery with DEM-derived slope and aspect. Monopoli et al.~\citep{monopoli2024} introduced BBUnet, which combines temporal optical imagery with topographic features for improved landslide mapping. Hafner et al.~\citep{tc-16-3517-2022} adapted DeepLabV3+ with deformable convolutions for avalanche detection in SPOT-6/7 imagery, explicitly incorporating DEM information. Similarly, Bianchi et al.~\citep{9254089} applied a U-Net architecture to Sentinel-1 SAR avalanche detection by combining backscatter differences with topographic features, reporting improvements over traditional signal-processing baselines. Together, these studies reinforce the prevailing assumption that terrain-aware priors are necessary to achieve accurate avalanche detection from satellite imagery.

However, it remains unclear whether auxiliary terrain information is genuinely necessary for reliable avalanche detection from SAR imagery. Although terrain-aware priors are widely assumed to be beneficial, modern deep networks may already be capable of learning the relevant spatial and radiometric structure directly from bi-temporal SAR inputs. This distinction is important in practice: if auxiliary inputs do not provide a consistent benefit, omitting them simplifies preprocessing, reduces model complexity, and improves scalability for large-area deployment. For this reason, avalanche mapping is investigated here as a unimodal SAR change-detection problem and systematically compared with multimodal alternatives that incorporate terrain-derived predictors.

\section{Material and Methods}
\label{sec:method}

\begin{figure*}[t]
\centering
\includegraphics[width=\linewidth]{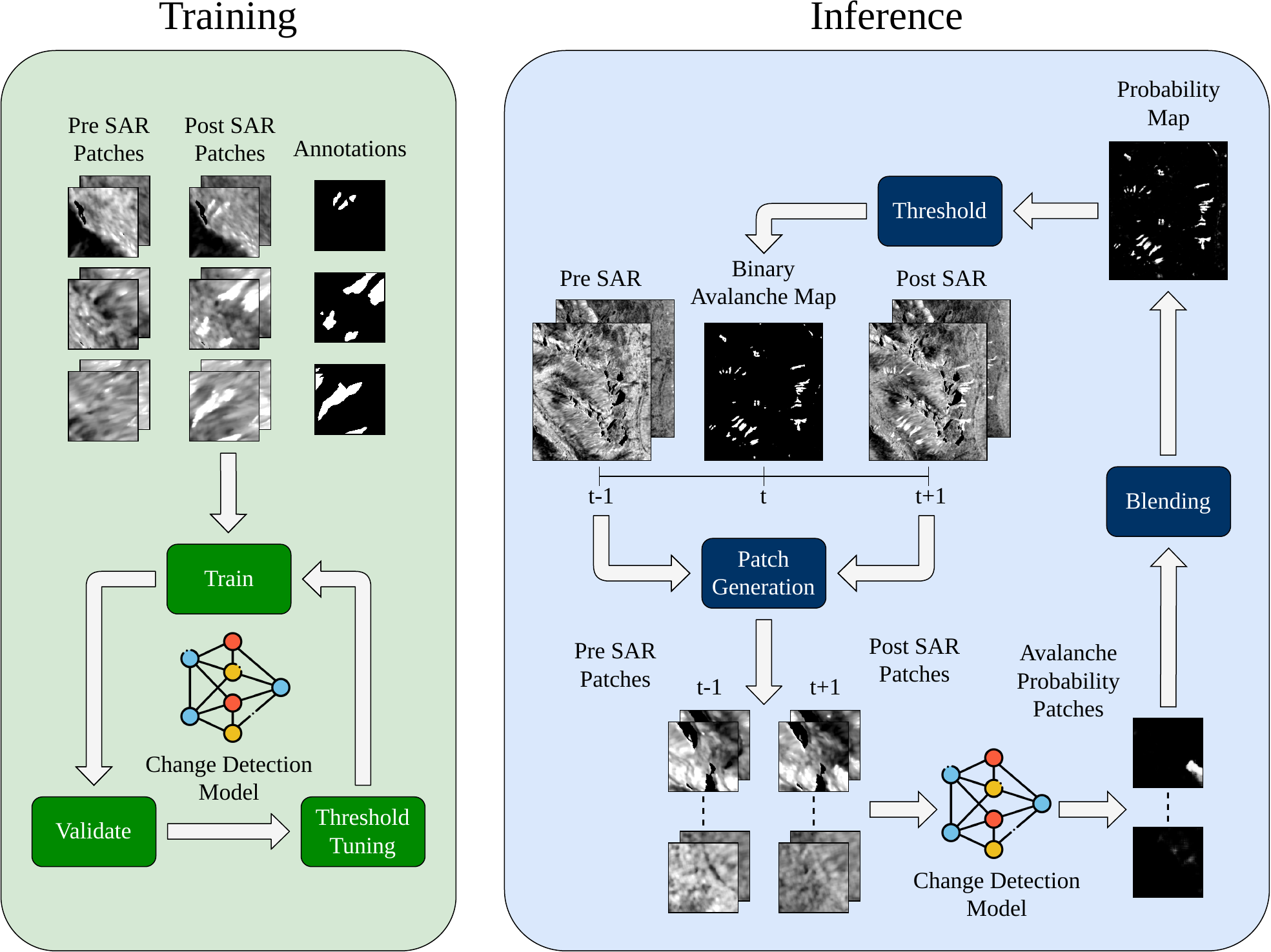}
\caption{Overview of the proposed avalanche mapping workflow. SAR images (VH and VV) are preprocessed (not shown). In the training stage (left), patches and annotations are used to fit a change detection model, followed by validation and threshold tuning. During inference (right), patches are extracted from pre- and post-event SAR images and passed through the model to obtain avalanche probability maps, which are then blended into a continuous map and thresholded to produce the final binary avalanche map.}
\label{fig:workflow}
\end{figure*}

This section describes the avalanche-mapping pipeline (\Cref{fig:workflow}), including Sentinel-1 preprocessing, dataset construction, model design, and large-scale inference with overlapping-tile blending.

\subsection{Data Preprocessing}

All Synthetic Aperture Radar (SAR) data used in this study were acquired by the Sentinel-1 satellite platform, which operates a C-band SAR sensor. It provides a swath width of approximately $250~\text{km}$, with a spatial resolution of $5~\text{m}$ in range and $20~\text{m}$ in azimuth. Freely available Level-1 Ground Range Detected (GRD) products were downloaded from the Copernicus Open Access Hub~\citep{copernicus_scihub}.

Data preprocessing was conducted using the Sentinel Application Platform (SNAP) software~\citep{esa_snap}, following a standard processing chain. This included application of the precise orbit files, radiometric calibration, geometric and radiometric terrain correction, speckle filtering, conversion to decibel (dB) scale and computation of the LIA raster for each scene. Pixel spacing was left to SNAP’s default Automatic setting, which preserves the native ground range resolution of Sentinel-1 IW GRD data ($\approx 10 \times 10~\text{m}$). For acquisitions over Nuuk, the effective spacing is approximately $5 \times 5~\text{m}$ due to the higher incidence angle and ground projection at polar latitudes. Dual-polarization channels VV and VH were available for all scenes, except for the Nuuk scene, which featured HH and HV polarizations. All VV--VH--LIA (or HH--HV--LIA) triplets were resampled to ensure they share the same width and height, allowing for consistent spatial alignment across input channels.

A $30~\text{m}$-resolution Digital Elevation Model (DEM) from the Copernicus programme~\citep{copernicus_dem} was used for SAR data preprocessing, as well as for computing the local incidence angle (LIA) and terrain slope. Because the DEM resolution is three times coarser than that of the Sentinel-1 imagery, the DEM is first upsampled using bilinear interpolation to match the spatial resolution of the SAR data, ensuring proper spatial alignment across modalities.

\subsection{Dataset Creation}
\begin{figure*}[t]
    \centering
    \includegraphics[width=\linewidth]{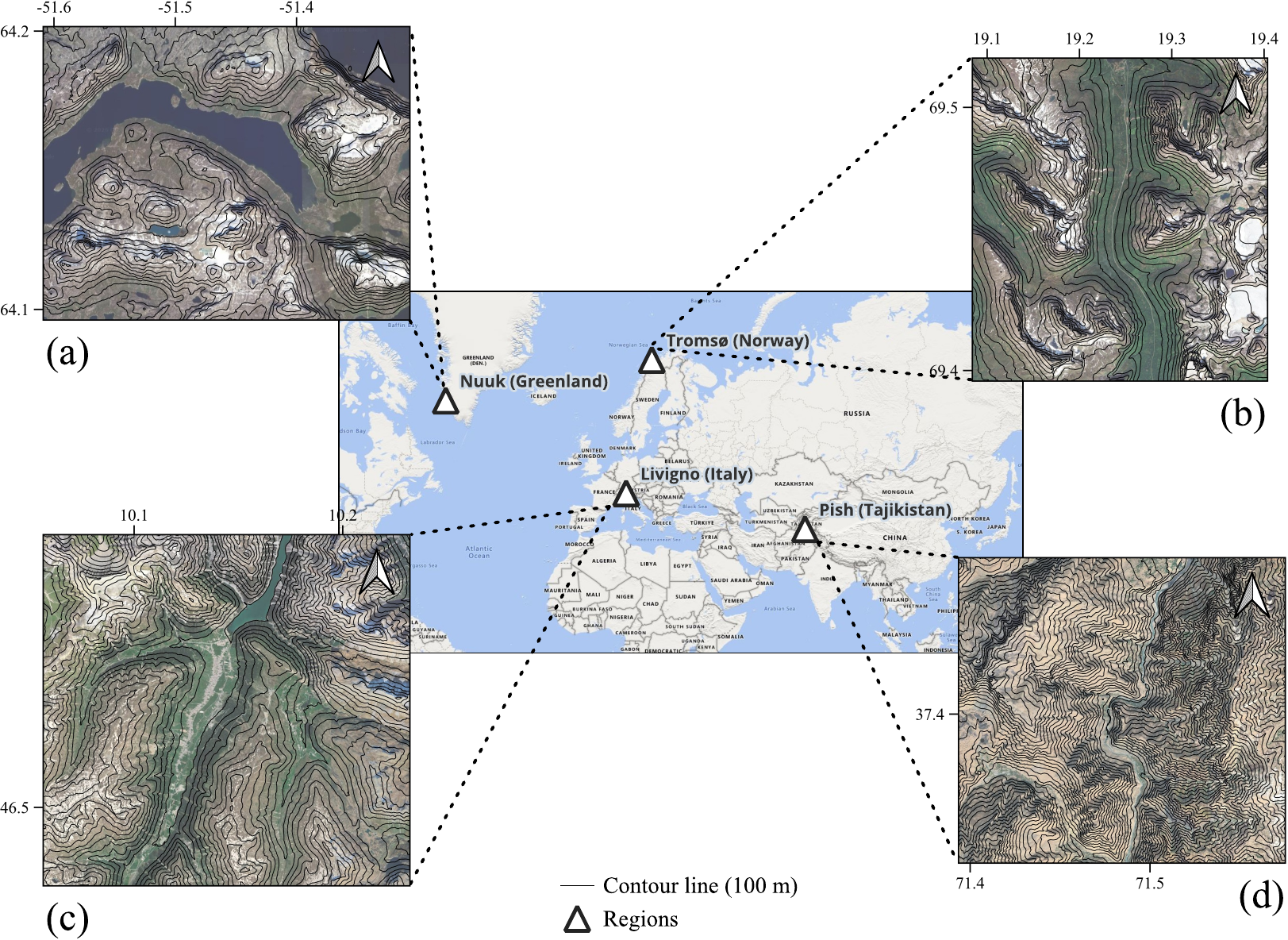}
    \caption{Geographic location of the study areas used in this work: Nuuk (a), Tromsø (b), Livigno (c), and Pish (d).}
    \label{fig:data-locations}
\end{figure*}

The dataset combines avalanche inventories from four distinct geographic regions: Livigno (Italy), Nuuk (Greenland), Pish (Shughnon District, Tajikistan), and Tromsø (Norway). These names serve only as geographic anchors; some mapped avalanches extend beyond the administrative or colloquial boundaries of the listed regions. The inventory size (polygon count) and spatial footprint for each region are summarized in \Cref{tab:aval-inventories}. The geographic distribution of the study areas is shown in \Cref{fig:data-locations}.

\begin{table}[t]
\centering
\begin{tabular}{lccc}
\toprule
\textbf{Event} & \textbf{Date} & \textbf{Extent (km$^{2}$)} & \textbf{Polygons (n)} \\
\midrule
Livigno  & 2024-04-03 & 509 & 231 \\
         & 2025-01-29 & 251 & 207 \\
         & 2025-03-18 & 488 &  79 \\
Nuuk     & 2016-04-13 & 226 & 117 \\
         & 2021-04-11 & 255 & 129 \\
Pish     & 2023-02-21 & 253 & 113 \\
Tromsø   & 2024-12-20 & 202 & 117 \\
\bottomrule
\end{tabular}
\caption{Avalanche inventories analyzed in this study. \textit{Date} is the post‑event SAR acquisition. \textit{Extent} is the planimetric area of the clipped analysis polygon (not the raster extent). \textit{Polygons} is the number of mapped avalanches.}
\label{tab:aval-inventories}
\end{table}

The selected event dates correspond to periods of high avalanche activity and were chosen based on the availability of extensive ground-truth evidence, including field observations, avalanche reports, and numerous site photographs acquired shortly after each event. These complementary sources enabled more complete and reliable delineation than visual interpretation of Sentinel-1 imagery alone.

Although the dataset spans four geographically distinct regions, it remains modest in size compared with large remote sensing benchmarks and may still reflect regional biases related to terrain, snow conditions, and radar geometry. To partially assess generalization, Tromsø was excluded from training and validation and used only for testing.

\begin{figure*}[t]
  \centering
  \begin{subfigure}[b]{0.13\linewidth}
    \centering
    \fbox{\includegraphics[width=\linewidth]{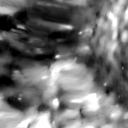}}
    \caption{Pre VV}
  \end{subfigure}
  \hfill
  \begin{subfigure}[b]{0.13\linewidth}
    \centering
    \fbox{\includegraphics[width=\linewidth]{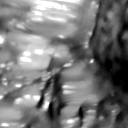}}
    \caption{Pre VH}
  \end{subfigure}
  \hfill
  \begin{subfigure}[b]{0.13\linewidth}
    \centering
    \fbox{\includegraphics[width=\linewidth]{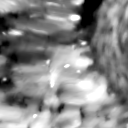}}
    \caption{Post VV}
  \end{subfigure}
  \hfill
  \begin{subfigure}[b]{0.13\linewidth}
    \centering
    \fbox{\includegraphics[width=\linewidth]{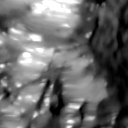}}
    \caption{Post VH}
  \end{subfigure}
  \hfill
  \begin{subfigure}[b]{0.13\linewidth}
    \centering
    \fbox{\includegraphics[width=\linewidth]{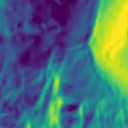}}
    \caption{LIA}
  \end{subfigure}
  \hfill
  \begin{subfigure}[b]{0.13\linewidth}
    \centering
    \fbox{\includegraphics[width=\linewidth]{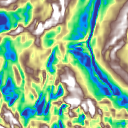}}
    \caption{Slope}
  \end{subfigure}
  \hfill
  \begin{subfigure}[b]{0.13\linewidth}
    \centering
    \fbox{\includegraphics[width=\linewidth]{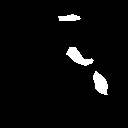}}
    \caption{Mask}
  \end{subfigure}

  \caption{Example of a $128 \times 128$ input patch used for avalanche mapping. From left to right, the panels show the four SAR backscatter channels, the resampled local incidence angle (LIA) and terrain slope, and the ground-truth segmentation mask. SAR channels are independently normalized using a $2$--$98\%$ percentile stretch. LIA and slope are visualized using the \texttt{viridis} and \texttt{terrain} colormaps, respectively.}
  \label{fig:avalanche-sample}
\end{figure*}

From each SAR scene, overlapping patches of size $32 \times 32$, $64 \times 64$, and $128 \times 128$ pixels, extracted with strides of 16, 32, and 64 pixels, respectively, are generated together with local incidence angle and slope. Each patch contains pre- and post-event VV/VH channels, LIA, slope, and a binary avalanche mask (see \Cref{fig:avalanche-sample}).

\setlength\fboxsep{0pt}
\setlength{\fboxrule}{0.4pt}

Overlapping patches enlarge the pool of positive samples and expose the model to the same regions in varying spatial contexts, promoting robustness to local variation and boundary effects; they also enable the inference-time blending that stitches tile predictions into seamless maps. All negative patches are retained. Although this results in a training set with a strong class imbalance, the distribution is rebalanced during training using a sampling strategy that ensures equal representation of positive and negative samples per epoch. This allows the model to benefit from the diversity of negative examples while maintaining balanced supervision; further implementation details are provided in the experimental section. The test split is constructed with equal numbers of positive and negative patches for each patch size. The resulting counts are summarized in \Cref{tab:aval-patches}.

\begin{table*}[t]
\centering
\begin{tabular}{l|cc|cc|cc}
\toprule
\multirow{3}{*}{\textbf{Event}} & \multicolumn{6}{c}{\textbf{Patch size}} \\
\cmidrule(lr){2-7}
 & \multicolumn{2}{c|}{\textbf{32}} & \multicolumn{2}{c|}{\textbf{64}} & \multicolumn{2}{c}{\textbf{128}} \\
 & \textbf{Pos.} & \textbf{Neg.} & \textbf{Pos.} & \textbf{Neg.} & \textbf{Pos.} & \textbf{Neg.} \\
\midrule
Livigno\_20240403 & 1353 & 13445 & 739 & 2919 & 399 & 480 \\
Livigno\_20250129 &  934 & 12460 & 561 & 2697 & 299 & 494 \\
Livigno\_20250318 &  439 & 25785 & 276 & 6174 & 162 & 1399 \\
Nuuk\_20160413    & 1366 & 30922 & 695 & 7265 & 366 & 1592 \\
Nuuk\_20210411    & 1458 & 34858 & 709 & 8286 & 367 & 1854 \\
Pish\_20230221    &  895 & 12599 & 491 & 2844 & 261 & 523 \\
Tromso\_20241220  &  907 &  8161 & 484 & 1727 & 261 & 262 \\
\midrule
\textbf{Total}    & \textbf{7352} & \textbf{138230} & \textbf{3955} & \textbf{31912} & \textbf{2115} & \textbf{6604} \\
\bottomrule
\end{tabular}
\caption{Number of extracted image patches for each avalanche event, reported separately for positive (Pos.) and negative (Neg.) samples across different patch sizes (32, 64, and 128 pixels).}
\label{tab:aval-patches}
\end{table*}

Due to the physical characteristics of radar imaging, terrain configurations such as steep slopes or complex topography can lead to signal loss, known as radar shadow or layover, which results in little or no backscatter reaching the satellite sensor. In such cases, measured values can fall far below the typical dynamic range, often below \SI{-40}{dB}. Additionally, preprocessing artifacts or corrupted pixels may produce non-finite values or extreme outliers.

To standardize inputs while retaining affected patches, SAR values are considered valid only within the range \SI{-40}{dB} to \SI{20}{dB}. All non-finite values and values outside this interval are treated as invalid. For each SAR channel, normalization is performed using the dataset-wide mean and standard deviation computed exclusively over valid observations. After normalization, invalid entries are replaced with a channel-specific sentinel value defined as the normalized value corresponding to \SI{-50}{dB},
\begin{equation}
s_c = \frac{-50 - \mu_c}{\sigma_c},
\end{equation}
where \(\mu_c\) and \(\sigma_c\) denote the mean and standard deviation of valid observations for channel \(c\). The value \SI{-50}{dB} lies well below the valid SAR range and is used only to define a sentinel that falls outside the normalized distribution of valid observations. This provides a consistent numerical marker for invalid measurements, including radar shadow, layover, or corrupted pixels, allowing the model to distinguish them during training and inference without requiring an additional masking channel.

Explicit masking of radar shadow and layover was also considered, but sentinel replacement was preferred to preserve affected patches and keep the input structure unchanged. Because these artifacts are usually spatially consistent between pre- and post-event acquisitions, their influence on change detection is partly reduced.

\subsection{Model}
A multi-branch Swin Transformer~V2~\citep{liu2022swin} in the tiny variant was implemented to explore multimodal fusion. The Swin Transformer was selected for its hierarchical representation and shifted-window attention mechanism, which enable efficient modeling of long-range spatial dependencies while maintaining computational efficiency. These characteristics are particularly relevant for avalanche detection and other remote sensing tasks, where contextual terrain patterns and spatial continuity play an important role. The architecture consists of two identical encoders with shared weights that process pre- and post-event SAR images, and a separate encoder that processes auxiliary inputs, namely local incidence angle (LIA) and terrain slope.

The encoded SAR representations are compared by computing the element-wise difference between the post-event and pre-event feature maps at the deepest encoder stage, producing a representation of temporal change. These change features are concatenated with topographic features extracted from the auxiliary branch.

The joint representation is then processed by a convolutional fusion module with residual connections, enabling joint spatio-temporal and terrain-aware reasoning. Following the dual-stream design proposed in~\citep{lu2023dual}, fusion is performed at the deep feature level, allowing the decoder to effectively integrate dynamic SAR-based changes with terrain-informed context. The fused representation is finally passed to a shared decoder that produces the binary avalanche segmentation mask.

\begin{figure*}
    \centering    \includegraphics[width=0.8\linewidth]{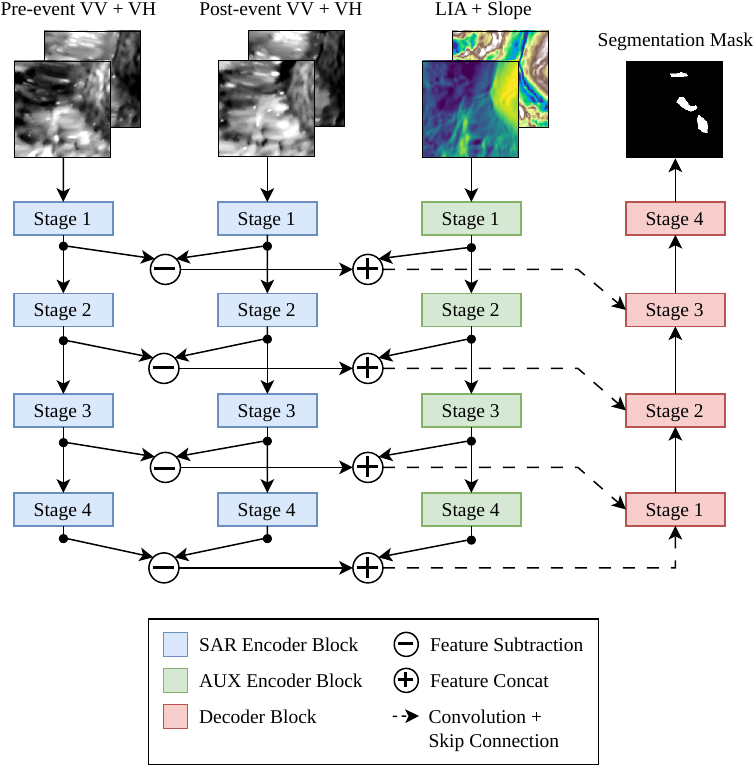}
    \vspace{8px}
    \caption{3D schematic of the proposed architecture. Pre- and post-event SAR images are processed by weight-sharing Swin Transformer encoders to extract temporally aligned features. Auxiliary topographic inputs (slope and local incidence angle) are processed by a separate Swin encoder. The element-wise difference between the deepest pre- and post-event SAR feature maps is fused with topographic features via a residual convolutional module and decoded by a hierarchical Swin Transformer, which progressively upsamples and refines the representation to produce a binary avalanche segmentation mask. This design enables joint reasoning over spatial, temporal, and topographic cues.}
    \label{fig:architecture}
\end{figure*}

Incorporating LIA and terrain slope data can provide useful spatial priors, particularly in complex alpine environments where local incidence angle varies sharply due to rugged terrain, affecting the radiometric response of avalanche deposits. Terrain slope is also a key physical factor in avalanche formation, making it a potentially informative cue for detection. The complete multimodal architecture contains 70.58 million trainable parameters and is illustrated in \Cref{fig:architecture}.

Ablation experiments were used to assess the contribution of auxiliary inputs and model complexity; the corresponding results are reported in \Cref{tab:patch-comparison,tab:patch-comparison-woaux}.

\paragraph{Training objective}
The model is trained using a weighted binary cross-entropy loss to account for the strong pixel-level imbalance between avalanche and background classes. The loss is defined as

\begin{equation}
\mathcal{L} = - w_p \, y \log(\hat{y}) - (1-y)\log(1-\hat{y}),
\end{equation}
where \(y\) denotes the ground-truth label, \(\hat{y}\) is the predicted probability, and \(w_p\) represents the weight assigned to the positive class. A value of \(w_p = 3.0\) is used to increase the penalty for misclassified avalanche pixels and reduce the dominance of background observations during training.

\subsection{Blending Logic}
\label{sec:blending}

When applying segmentation models to large remote sensing images, inputs are typically divided into smaller overlapping tiles to fit GPU memory constraints. Overlapping tiles ensure that border pixels receive sufficient spatial context, but they also produce multiple predictions for the same spatial locations. A blending strategy is therefore required to combine these predictions into a single coherent output.

In this study, several tile fusion strategies were evaluated to assess their effect on segmentation consistency across tile boundaries. The investigated methods include per-pixel averaging, max blending, min blending, Gaussian-weighted blending, and center cropping.

\paragraph{Per-pixel averaging}
A common method computes the mean prediction for each overlapping pixel:

\begin{equation}
\hat{P}(x, y) = \frac{1}{N(x, y)} \sum_{i=1}^{N(x, y)} P_i(x, y)
\end{equation}
where \(P_i(x,y)\) is the prediction from the \(i\)-th tile at pixel \((x,y)\), and \(N(x,y)\) is the number of tiles covering that location. Averaging smooths transitions between tiles and reduces boundary artifacts.

\paragraph{Max and min blending}
Max blending selects the maximum predicted value across tiles,

\begin{equation}
\hat{P}(x, y) = \max_{i=1}^{N(x, y)} P_i(x, y)
\end{equation}
while min blending selects the minimum,

\begin{equation}
\hat{P}(x, y) = \min_{i=1}^{N(x, y)} P_i(x, y)
\end{equation}
These strategies represent sensitivity-oriented and conservative fusion modes, respectively.

\paragraph{Gaussian-weighted blending}
To reduce the influence of tile borders, spatial weighting can be applied:

\begin{equation}
\hat{P}(x, y) = \frac{\sum_{i=1}^{N(x, y)} W_i(x, y) P_i(x, y)}{\sum_{i=1}^{N(x, y)} W_i(x, y)}
\end{equation}
where \(W_i(x,y)\) is a spatial weighting function assigning higher importance to central tile pixels.

\paragraph{Center cropping}
Finally, center cropping discards tile borders and retains only the central region \(C_i\):

\begin{equation}
\hat{P}(x, y) = P_i(x, y), \quad \text{if } (x, y) \in C_i
\end{equation}
This approach avoids blending entirely and relies on the most reliable tile regions.

All blending strategies described above are evaluated in the experiments to assess how different fusion assumptions influence prediction consistency in overlapping regions.

\section{Experimental Setup}
\label{sec:experiments}
This section summarizes the training, validation, and evaluation setup, including data splits, augmentation, baseline comparisons, threshold selection, and full-scene inference.

\subsection{Data Splits and Training Configuration}
The model is trained and evaluated from avalanche events in Livigno, Nuuk, and Pish. A subset of these samples was reserved as a validation set to monitor performance during training, tune hyperparameters, and prevent overfitting. Importantly, avalanche data from Tromsø was excluded from both training and validation and was held out entirely as an independent test set, ensuring that the evaluation reflects the model’s ability to generalize to a geographically distinct region.

Training employs the AdamW optimizer with a learning rate of \(1 \times 10^{-4}\), combined with a two-stage learning rate schedule consisting of a linear warm-up for the first 10 epochs, followed by cosine annealing over the remaining 100 epochs. To evaluate the impact of input resolution, three separate training runs are performed, each using a fixed patch size of $32 \times 32$, $64 \times 64$, or $128 \times 128$.

\subsection{Handling of Class Imbalance and Data Augmentation}
Avalanche pixels are strongly underrepresented relative to background pixels. To mitigate this imbalance at the patch level, a custom sampling strategy is employed during training. Each training epoch is constructed using a balanced sampler that selects an equal number of positive samples (patches containing at least one avalanche pixel) and negative samples. 

To ensure fair representation across different events, negative samples are drawn uniformly from each event’s pool. This event-aware balancing encourages the model to generalize across varying geographic and temporal contexts.

In addition to balancing strategies, geometric and radiometric data augmentations are applied during training to further enhance model robustness and generalization. Geometric augmentations include random horizontal flipping, 90-degree rotations, and mild affine transformations (rotation, translation, scaling, and shearing) to simulate small geolocation inaccuracies and increase spatial variability. These transformations are applied consistently to pre-event, post-event, auxiliary, and mask tensors to preserve spatial alignment. Radiometric augmentations are designed to reflect natural variability in SAR backscatter. These include the addition of Gaussian noise and small random gain perturbations to the normalized SAR channels, mimicking real-world variations such as sensor noise and scene-dependent reflectance fluctuations. Together, these augmentations improve the model’s ability to handle diverse terrain configurations and signal inconsistencies encountered during inference.

\subsection{Effect of Auxiliary Inputs and Baseline Comparison}
To evaluate the contribution of the auxiliary input (LIA and terrain slope), an experiment was conducted in which models were trained both with and without the auxiliary channels. This design enabled quantification of the extent to which auxiliary information enhances avalanche detection performance beyond what can be achieved using pre- and post-event SAR imagery alone. In addition, experiments included comparisons with established change detection baselines, ensuring that performance gains can be contextualized against widely used methods and allowing assessment of whether avalanche-specific modeling provides benefits beyond generic change detection approaches.

\subsection{Evaluation Metrics and Threshold Tuning}
During training, model performance is monitored on the validation set using a comprehensive set of metrics, including F1-score, precision, recall, intersection-over-union (IoU), average precision (AUPRC), and average validation loss. To select the optimal decision threshold, the F1-score is evaluated across the unique set of predicted probabilities from the validation set. Each score is treated as a candidate threshold, binarizing the output to compute precision, recall, F1 and IoU. This strategy adapts threshold selection to the model’s confidence distribution, yielding a more precise decision boundary. The threshold that maximizes the F1-score is used for testing and inference, as it provides the best overall agreement with the ground truth. Final model selection is based on the highest AUPRC on the validation set, with its corresponding threshold retained for deployment.

Additionally, recall-oriented thresholds are optimized using the F2-score ($\beta = 2$) to maximize recall and analyze precision-recall trade-offs, thereby assessing the sensitivity of avalanche detection performance to threshold selection.

\subsection{Post-processing and Morphological Refinement}

To suppress isolated false detections and fill small gaps within predicted avalanche segments, a morphological closing operation is applied to the binary mask obtained after thresholding. The closing consists of a dilation followed by an erosion, both using a $3 \times 3$ structuring element and applied for one iteration. This simple post-processing step helps remove small holes and connect fragmented detections, thereby improving the spatial coherence of the final avalanche maps while preserving the overall extent of predicted regions. The structuring element was intentionally kept small to limit the influence of the operation on the size of detected avalanche areas.

\subsection{Blending and Large-Scale Inference}
The model is first evaluated on individual patches to obtain an initial performance estimate, and subsequently tested over the full region using the blending strategies described in \Cref{sec:blending}. This evaluation examines how different blending schemes and threshold-tuning choices, independent of architectural design, influence the final results.

For large-scale inference, the study adopts a tiling-based approach in which the input scene is divided into overlapping patches to fit GPU memory constraints. Each patch prediction is merged using the selected blending strategy to form a continuous probability map, which is subsequently thresholded based on the optimal threshold obtained during validation. This setup mirrors how avalanche detection would operate in practice, ensuring that performance measured on small patches reflects the model’s behavior when applied to large regional scenes.

\section{Results and Discussion}
\label{sec:results}
This section presents the empirical evaluation of the model. It examines generalization to an unseen region, the influence of patch size, and the effect of auxiliary inputs. The results are compared against established change-detection baselines, and thresholding behavior is analyzed under both F1- and F2-based tuning. Full-scene inference experiments assess blending strategies and object-level detection completeness, followed by analyzes of size-class performance, computational efficiency, and key limitations.

\subsection{Generalization and Patch-Size Sensitivity}
Generalization was evaluated on an independent test scene over Tromsø, fully excluded from training and validation. The scene is tiled with 50\% overlap (see \Cref{tab:aval-patches}), ensuring unbiased evaluation across unseen terrain.

Although Swin Transformer V2 Tiny was originally designed for $256 \times 256$ inputs, experiments with a truncated three-stage variant produced virtually identical results to the full four-stage model. The final encoder stage was therefore omitted, reducing the parameter count from 70.58~M to 4.01~M while preserving performance. \Cref{fig:architecture} illustrates the full four-stage design, but the implementation used in this study employs the reduced variant.

\Cref{tab:patch-comparison} reports the results for different patch sizes. In these experiments, morphological erosion and dilation were applied at the patch level, i.e., each individual patch prediction was post-processed before computing the metrics. The proposed model trained with $128 \times 128$ patches yielded the highest F1 (0.7971) and IoU (0.6608). The results indicate that larger patches provide slightly more context for delineating avalanche runouts, although performance differences remain minor, indicating robustness to patch size. Practically, larger patches are more efficient, covering wider areas per forward pass while maintaining comparable or marginally improved accuracy.

\subsection{Influence of Auxiliary Inputs}
The contribution of the auxiliary input branch was evaluated in a separate experiment reported in \Cref{tab:patch-comparison-woaux}. Results showed that removing the auxiliary branch led to virtually identical, and in some cases slightly improved, performance, suggesting that the additional inputs did not provide a consistent benefit. Further experiments with only one of the two auxiliary channels individually are not reported in detail, as they yielded the same outcome, indicating that neither LIA nor slope provided a meaningful benefit for avalanche detection in this setting. These findings suggest that, although such inputs can simplify traditional avalanche detection workflows, comparable performance can be achieved using SAR imagery alone in this setting.

Unless otherwise stated, all subsequent references to the proposed model correspond to the variant without the auxiliary branch. This simplification reduces the parameter count from 4.01~M to 2.39~M, representing a reduction of about 40\% compared to the configuration with the auxiliary branch and without the fourth stage.

\begin{table}
\centering
\begin{tabular}{lcccc}
\toprule
\textbf{Patch size} & \textbf{Recall} & \textbf{Precision} & \textbf{F1-score} & \textbf{IoU} \\
\midrule
$32 \times 32$   & 0.7553 & 0.8014 & 0.7777 & 0.6362 \\
$64 \times 64$   & 0.7408 & 0.8380 & 0.7864 & 0.6480 \\
$128 \times 128$ & 0.7854 & 0.8092 & \textbf{0.7971} & \textbf{0.6608} \\
\bottomrule
\end{tabular}
\caption{Performance metrics on the test set for different patch sizes with auxiliary input. Bold values indicate the best F1 and IoU scores, which were the metrics optimized during threshold tuning.}
\label{tab:patch-comparison}
\end{table}

\begin{table}
\centering
\begin{tabular}{lcccc}
\toprule
\textbf{Patch size} & \textbf{Recall} & \textbf{Precision} & \textbf{F1-score} & \textbf{IoU} \\
\midrule
$32 \times 32$  & 0.7566 & 0.8157 & 0.7850 & 0.6461 \\
$64 \times 64$   & 0.7635 & 0.8148 & 0.7880 & 0.6504 \\
$128 \times 128$  & 0.7925 & 0.8131 & \textbf{0.8027} & \textbf{0.6704} \\
\bottomrule
\end{tabular}
\caption{Performance metrics on the test set for different patch sizes without auxiliary input. Bold values indicate the best F1 and IoU scores, which were the metrics optimized during threshold tuning.}
\label{tab:patch-comparison-woaux}
\end{table}

Bianchi et al.~\citep{9254089} reported a modest improvement of about 0.3\% in detection accuracy when incorporating slope information, and an additional gain of 1.4\% when including the Potential Avalanche Release (PAR) map as an attention mask. In their approach, the PAR constrains the model to ignore terrain that is physically implausible for avalanche initiation. In contrast, the proposed alternative applies such physical constraints after inference, once all patch predictions are merged into a continuous probability map, removing the need to provide additional spatial cues during training. By running avalanche simulations under extreme conditions, physically implausible terrain can be excluded in post-processing, ensuring realistic spatial filtering while preserving model generality.

\subsection{F1-Threshold Analysis}
Since threshold tuning in validation was driven solely by maximizing F1 (and, by extension, IoU), any apparent advantage in precision or recall across patch sizes is incidental and should not be interpreted as a genuine performance difference. Validation-derived thresholds generalized well and maintained a balanced precision-recall trade-off. Precision consistently exceeded recall, largely due to boundary ambiguity rather than missed events: at Sentinel-1’s $\approx 10$ m resolution, runout edges are often diffuse, and small shifts in predicted footprints may appear as false positives or false negatives. Qualitative examples in \Cref{fig:qualitative} support this interpretation, with most discrepancies occurring near uncertain boundaries or involving slightly undersized predictions.

\newcommand{\colsep}{0.02\linewidth}
\newcommand{\colw}{0.144\linewidth} 

\begin{figure*}[t]
\centering

{\small
\makebox[\textwidth][c]{
\makebox[\colw][c]{VV Composite}\hspace{\colsep}
\makebox[\colw][c]{VH Composite}\hspace{\colsep}
\makebox[\colw][c]{Pred. Prob.}\hspace{\colsep}
\makebox[\colw][c]{Pred. Mask}\hspace{\colsep}
\makebox[\colw][c]{Ground Truth}
}
}

\vspace{-6pt}

\makebox[\textwidth][c]{
\begin{subfigure}[t]{\colw}
\centering
\fbox{\includegraphics[width=\linewidth]{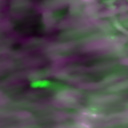}}
\end{subfigure}\hspace{\colsep}
\begin{subfigure}[t]{\colw}
\centering
\fbox{\includegraphics[width=\linewidth]{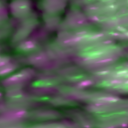}}
\end{subfigure}\hspace{\colsep}
\begin{subfigure}[t]{\colw}
\centering
\fbox{\includegraphics[width=\linewidth]{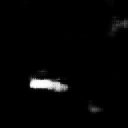}}
\end{subfigure}\hspace{\colsep}
\begin{subfigure}[t]{\colw}
\centering
\fbox{\includegraphics[width=\linewidth]{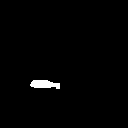}}
\end{subfigure}\hspace{\colsep}
\begin{subfigure}[t]{\colw}
\centering
\fbox{\includegraphics[width=\linewidth]{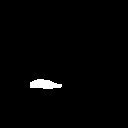}}
\end{subfigure}
}

\vspace{10pt}

\makebox[\textwidth][c]{
\begin{subfigure}[t]{\colw}
\centering
\fbox{\includegraphics[width=\linewidth]{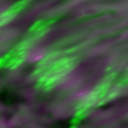}}
\end{subfigure}\hspace{\colsep}
\begin{subfigure}[t]{\colw}
\centering
\fbox{\includegraphics[width=\linewidth]{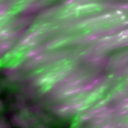}}
\end{subfigure}\hspace{\colsep}
\begin{subfigure}[t]{\colw}
\centering
\fbox{\includegraphics[width=\linewidth]{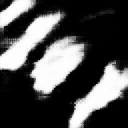}}
\end{subfigure}\hspace{\colsep}
\begin{subfigure}[t]{\colw}
\centering
\fbox{\includegraphics[width=\linewidth]{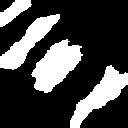}}
\end{subfigure}\hspace{\colsep}
\begin{subfigure}[t]{\colw}
\centering
\fbox{\includegraphics[width=\linewidth]{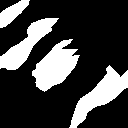}}
\end{subfigure}
}

\vspace{10pt}

\makebox[\textwidth][c]{
\begin{subfigure}[t]{\colw}
\centering
\fbox{\includegraphics[width=\linewidth]{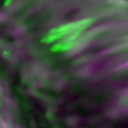}}
\end{subfigure}\hspace{\colsep}
\begin{subfigure}[t]{\colw}
\centering
\fbox{\includegraphics[width=\linewidth]{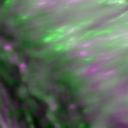}}
\end{subfigure}\hspace{\colsep}
\begin{subfigure}[t]{\colw}
\centering
\fbox{\includegraphics[width=\linewidth]{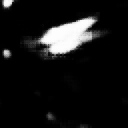}}
\end{subfigure}\hspace{\colsep}
\begin{subfigure}[t]{\colw}
\centering
\fbox{\includegraphics[width=\linewidth]{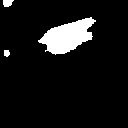}}
\end{subfigure}\hspace{\colsep}
\begin{subfigure}[t]{\colw}
\centering
\fbox{\includegraphics[width=\linewidth]{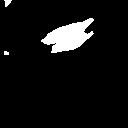}}
\end{subfigure}
}

\caption{Visual results for selected test patches. From left to right: VV RGB composite (pre-post-pre), VH RGB composite, predicted probability map, thresholded binary mask, and ground truth mask.}
\label{fig:qualitative}
\end{figure*}

Deep-learning methods for avalanche mapping in Sentinel-1 SAR imagery remain largely unexplored: only a small number of prior studies have been published, all based on proprietary datasets and heterogeneous annotation protocols. Reported F1-scores of 0.66 \citep{9254089} and 0.78 \citep{sinha:hal-02278230}, for instance, cannot be compared directly with the present results because they were obtained in different geographical settings, characterized by dissimilar snow regimes and terrain morphologies. To support rigorous, reproducible benchmarking, the annotated dataset is released with this paper, with access information provided in \Cref{sec:data_availability}.

\subsection{Baselines Comparison}

Several baseline models were originally developed for optical change detection rather than SAR imagery. Because SAR differs substantially from optical data in radiometry and noise characteristics, and because all models were trained from scratch without optical pretraining, these baselines should be interpreted as strong generic references rather than SAR-specialized methods.

To contextualize these results, \Cref{tab:baseline-comparison} presents a comparison with common change detection baselines on the $128 \times 128$ test set, including SiamUnet~\citep{daudt2018fully}, STANet~\citep{Chen2020}, BIT~\citep{chen2021remote}, SNUNet-CD~\citep{fang2021snunet}, ChangeFormer~\citep{bandara2022transformer}, TinyCD~\citep{codegoni2023tinycd}, and STNet~\citep{10219826}. The proposed avalanche-specific architecture consistently achieved higher F1 and IoU scores than these generic approaches. This improvement likely stems from the Swin Transformer’s ability to capture long-range spatial dependencies and hierarchical representations from SAR backscatter, which are particularly important for detecting subtle avalanche-induced changes.

Notably, TinyCD~\citep{codegoni2023tinycd}, despite its minimal size of only 0.29\,M parameters, achieved competitive results. However, its performance remained below that of the Swin-based architecture, highlighting the advantage of transformer-based models in capturing complex spatial patterns in SAR imagery.

\begin{table*}[t]
\centering
\begin{tabular}{lccc}
\toprule
\textbf{Model} & \textbf{Params (M)} & \textbf{F1} & \textbf{IoU} \\
\midrule
SiamUnet-diff~\citep{daudt2018fully}      & 1.35  & 0.7370 & 0.5836 \\
SiamUnet-conc~\citep{daudt2018fully}      & 1.35  & 0.7498 & 0.5997 \\
STANet~\citep{Chen2020}                     & 2.42  & 0.6460 & 0.4771 \\
BIT~\citep{chen2021remote}                  & 2.97  & 0.6999 & 0.5383 \\
SNUNet-CD~\citep{fang2021snunet}            & 12.03 & 0.7546 & 0.6059 \\
ChangeFormer~\citep{bandara2022transformer} & 55.26 & 0.7720 & 0.6197 \\
TinyCD~\citep{codegoni2023tinycd}           & 0.29  & 0.7660 & 0.6208 \\
STNet~\citep{10219826}                      & 14.60 & 0.7777 & 0.6362 \\
Swin-UNet (this work)                      & 2.39 & \textbf{0.8027} & \textbf{0.6608} \\
\bottomrule
\end{tabular}
\caption{Comparison of change detection models on the test set. 
Only F1 and IoU are reported, as F1 was used as the primary optimization target during validation. 
Bold indicates the best results. Parameter counts are in millions (M). Models are listed by publication date.}
\label{tab:baseline-comparison}
\end{table*}

\subsection{Performance under Full-Scene Inference Conditions}

To provide a more interpretable object-level evaluation, avalanche polygons are analyzed according to the European Avalanche Warning Services (EAWS) size classification~\citep{eaws_standards}, summarized in \Cref{tab:eaws-sizes}. This classification groups avalanches by their typical release volume and associated impact.

\begin{table}
\centering
\begin{tabular}{cll}
\toprule
\textbf{Size} & \textbf{Description} & \textbf{Typical Release Volume (m$^3$)} \\
\midrule
1 & Small & $<10^{2}$ \\
2 & Medium & $10^{2}$--$10^{3}$ \\
3 & Large & $10^{3}$--$10^{4}$ \\
4 & Very Large & $10^{4}$--$10^{5}$ \\
5 & Extremely large & $>10^{5}$ \\
\bottomrule
\end{tabular}
\caption{Avalanche size classification according to the European Avalanche Warning Services (EAWS) standards~\citep{eaws_standards}.}
\label{tab:eaws-sizes}
\end{table}

Each reference polygon in the Tromsø test inventory was manually annotated according to this size classification. The inventory contains 117 reference polygons in total, with no avalanches of size~5. Size~1 avalanches are excluded from the object-level evaluation, as their spatial extent can fall below the effective ground resolution of Sentinel-1 imagery, making reliable detection inherently difficult. Consequently, all polygon-level metrics reported in the following tables are computed over 112 reference polygons corresponding to size classes 2--4.

\Cref{tab:blending-comparison} compares blending strategies for full-scene inference. In addition to pixel-level metrics, a Hit~\% score is reported, counting a reference polygon as detected when at least 50\% of its area is predicted as avalanche. This object-level measure complements pixel-wise evaluation and is less sensitive to minor boundary inaccuracies.

\begin{table*}
\centering
\begin{tabular}{lccccc}
\toprule
\textbf{Mode} & \textbf{Rec.} & \textbf{Prec.} & \textbf{F1} & \textbf{IoU} & \textbf{Hit~\%} \\
\midrule
None         & 0.7892 & 0.8059 & 0.7975 & 0.6632 & 62.50\% (70/112) \\
Min          & 0.7710 & \textbf{0.8338} & 0.8012 & 0.6683 & 60.71\% (68/112) \\
Max          & \textbf{0.8136} & 0.7924 & 0.8029 & 0.6707 & \textbf{68.75\% (77/112)} \\
Mean         & 0.7887 & 0.8211 & 0.8046 & 0.6731 & 64.29\% (72/112) \\
Gaussian     & 0.7928 & 0.8199 & \textbf{0.8061} & \textbf{0.6752} & 65.18\% (73/112) \\
Center Crop  & 0.7948 & 0.8171 & 0.8058 & 0.6747 & 65.18\% (73/112) \\
\bottomrule
\end{tabular}
\caption{Performance metrics and total polygon hit rate. Best values per metric are shown in bold.}
\label{tab:blending-comparison}
\end{table*}

Each fusion rule introduces a distinct trade-off. The \textit{min} operator suppresses uncertain boundary pixels and delivers the highest precision, while the \textit{max} rule retains every positive response, maximizing recall and the avalanche hit rate. \textit{Gaussian} strategy provided the best overall agreement with the ground-truth masks, achieving the top F1 (0.8061) and IoU (0.6752) scores, although the results were nearly identical to those obtained with the \textit{Mean} and \textit{Center Crop} strategies. These overlapping strategies effectively suppress boundary noise and artifacts that arise during patch transitions. In contrast, the non-overlapping (\textit{None}) setting still achieved competitive metrics, showing that the model can reliably predict labels up to patch edges. Nevertheless, when mosaicking adjacent tiles, small misalignments can appear as visible seams, slightly reducing the perceived map quality even though quantitative scores remain high.

\subsection{F2-Thresholding for Large-Scale Inference}
The relatively limited hit rates observed across the blending strategies, together with qualitative inspection of missed avalanches, motivated an additional analysis using F2-based threshold tuning. This experiment is intended to approximate a recall-oriented operating regime that may be preferable in practical monitoring workflows, where the objective is to flag as many candidate avalanches as possible for subsequent review rather than to produce the most conservative map. The F2-score is defined as
\begin{equation}
F_{\beta} = (1 + \beta^{2}) \cdot \frac{\text{Precision} \cdot \text{Recall}}{(\beta^{2} \cdot \text{Precision}) + \text{Recall}},
\end{equation}
where $\beta$ controls the relative importance of recall to precision. Setting $\beta = 2$ gives
\begin{equation}
F_{2} = \frac{5 \cdot \text{Precision} \cdot \text{Recall}}{4 \cdot \text{Precision} + \text{Recall}},
\end{equation}
thereby assigning four times more weight to recall than to precision. In contrast, the F1-optimized threshold is more conservative and favors overall mask agreement. The comparison between these two regimes therefore helps clarify how threshold choice affects practical deployment: F1 is preferable when boundary quality and false-positive control are primary, whereas F2 is preferable when maximizing event coverage is more important.

\begin{figure}
  \centering  \includegraphics[width=\linewidth]{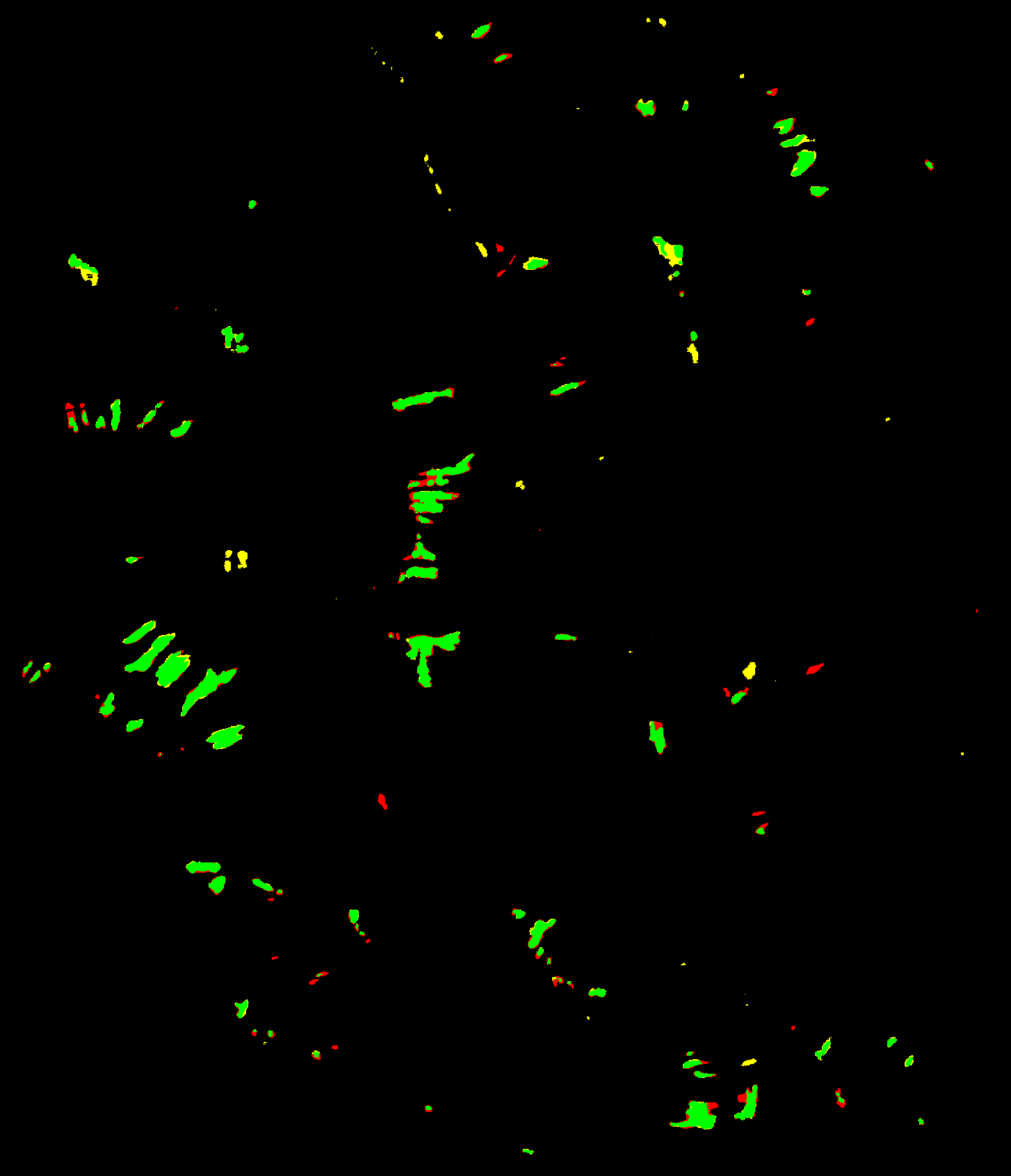}
{\small
  \makebox[0.90\linewidth][c]{
    \textcolor{black}{\rule{1.2em}{0.7em}}\,TN\hspace{1em}
    \textcolor{green}{\rule{1.2em}{0.7em}}\,TP\hspace{1em}
    \textcolor{red}{\rule{1.2em}{0.7em}}\,FN\hspace{1em}
    \textcolor{yellow}{\rule{1.2em}{0.7em}}\,FP
  }
  }
  \caption{Confusion map of predicted avalanches across the Tromsø region using the \textit{Gaussian} blending strategy, generated with the threshold optimized for the F1-score.}
  \label{fig:confusion_f1}
\end{figure}

\begin{figure}
  \centering  \includegraphics[width=\linewidth]{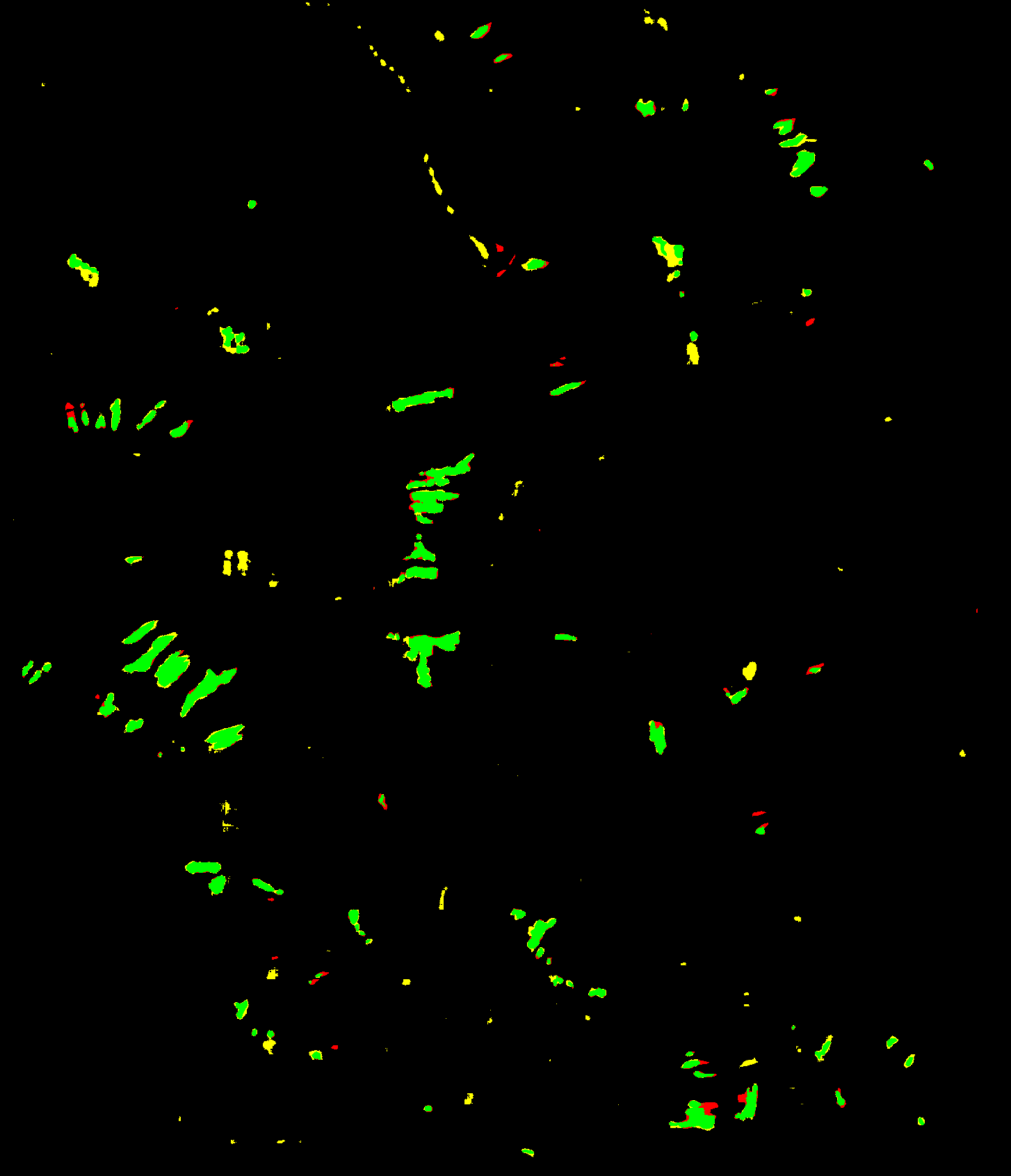}
{\small
  \makebox[0.90\linewidth][c]{
    \textcolor{black}{\rule{1.2em}{0.7em}}\,TN\hspace{1em}
    \textcolor{green}{\rule{1.2em}{0.7em}}\,TP\hspace{1em}
    \textcolor{red}{\rule{1.2em}{0.7em}}\,FN\hspace{1em}
    \textcolor{yellow}{\rule{1.2em}{0.7em}}\,FP
  }
  }
  \caption{Confusion map of predicted avalanches across the Tromsø region using the \textit{Max} blending strategy, generated with the threshold optimized for the F2-score.}
  \label{fig:confusion_f2}
\end{figure}

The impact of this change is illustrated in \Cref{fig:confusion_f1} and \Cref{fig:confusion_f2}, showing confusion maps for the F1-optimized Gaussian blending (maximizing overall agreement with the ground truth) and the F2-optimized max blending (capturing more avalanches through higher recall). As summarized in \Cref{tab:blend-f1f2}, the F2-tuned configuration substantially increases recall and overall avalanche coverage, resulting in higher hit rates across the test set. This improvement comes at the cost of lower precision, mainly due to additional false positives in snow accumulation zones and wind-redistributed areas.

Whether this trade-off is acceptable depends on the intended use case: for analyst-in-the-loop screening, such as regional post-event assessment by avalanche warning services (e.g., EAWS) or civil protection agencies, the increase in recall may justify the additional review effort, whereas fully automatic mapping may favor the more conservative F1-based setting. Operational requirements may differ across agencies and use cases; therefore, F1- and F2-based thresholds should be interpreted as two representative operating points rather than universally optimal settings.

\begin{table*}[t]
\centering
\begin{tabular}{lcccccc}
\toprule
\textbf{Mode} & \textbf{Rec.} & \textbf{Prec.} & \textbf{F1} & \textbf{F2} & \textbf{IoU} & \textbf{Hit~\%} \\
\midrule
Max (F1-opt)      & 0.8136 & 0.7924 & 0.8029 & 0.8076 & 0.6707 & 68.75\% (77) \\
Gauss (F1-opt)    & 0.7928 & \textbf{0.8199} & \textbf{0.8061} & 0.7991 & \textbf{0.6752} & 65.18\% (73) \\
Max (F2-opt)      & \textbf{0.8771} & 0.7021 & 0.7799 & \textbf{0.8414} & 0.6392 & \textbf{80.36\% (90)} \\
Gauss (F2-opt)    & 0.8633 & 0.7345 & 0.7937 & 0.8324 & 0.6579 & 78.57\% (88) \\
\bottomrule
\end{tabular}
\caption{Metrics and total polygon hit rate with threshold tuning (F1/F2). F2 emphasizes recall four times more than precision.}
\label{tab:blend-f1f2}
\end{table*}

As shown in \Cref{tab:hitrates-by-size-short}, the F2-based threshold markedly improves the detection of smaller avalanches, approximately doubling the hit rate for size~2 events compared to the F1-optimized models, while maintaining near-perfect detection for medium and large avalanches. This demonstrates, as expected, that F2 tuning effectively reduces the omission of small but physically relevant events that are often missed under the more conservative F1-optimized thresholds.

\begin{table*}[t]
\centering
\begin{tabular}{lccc}
\toprule
\textbf{Mode} & \textbf{Hit-Size 2 (\%)} & \textbf{Hit-Size 3 (\%)} & \textbf{Hit-Size 4 (\%)} \\
\midrule
Max (F1-opt)      & 32.00 (8/25)  & 74.65 (53/71) & 100.00 (16/16) \\
Gauss (F1-opt)    & 28.00 (7/25)  & 70.42 (50/71) & 100.00 (16/16) \\
Max (F2-opt)      & \textbf{64.00 (16/25)} & \textbf{81.69 (58/71)} & 100.00 (16/16) \\
Gauss (F2-opt)    & 60.00 (15/25) & 80.28 (57/71) & 100.00 (16/16) \\
\bottomrule
\end{tabular}
\caption{Avalanche polygon hit rates by EAWS size class (2–4).}
\label{tab:hitrates-by-size-short}
\end{table*}

\subsection{Computational Efficiency and Scalability}

Beyond evaluation accuracy, computational efficiency was also assessed. Using $128 \times 128$ patches with a stride of $64$, the system processes SAR scenes at roughly $87~\text{km}^2/\text{s}$ on an NVIDIA A100 GPU, meaning that the entire Alps ($\sim\!200{,}000~\text{km}^2$) can be processed in under 40 minutes under the tested hardware configuration. This runtime can be further reduced by approximately a factor of four if overlapping tiles and blending are disabled, although this may introduce boundary artifacts in the resulting maps.

These measurements should be interpreted as reference benchmarks obtained on high-end hardware; inference on less powerful GPUs remains feasible but proportionally slower, while training is substantially more computationally demanding.

\subsection{Limitations and Concluding Remarks}

Integrating auxiliary data during post-processing offers a promising way to enhance model performance. Detection performance is strongly influenced by snowpack evolution between acquisitions. In particular, snow melting occurring after avalanche release can reduce backscatter contrast, leading to missed detections, while widespread refreezing may produce strong backscatter increases over entire slopes, resulting in large false positives. Additional systematic errors are associated with anthropogenic surfaces such as ski slopes or frozen lakes, where temporal changes in surface roughness or dielectric properties can mimic avalanche-like signatures. 

These effects highlight that detection reliability is governed not only by model capacity, but also by the physical state of the snowpack at acquisition time. This helps explain the behavior under different thresholding strategies: conservative, F1-optimized thresholds tend to suppress many of these ambiguous responses, improving precision at the cost of missing marginal avalanches, whereas recall-oriented F2 tuning retains a larger fraction of these uncertain detections, increasing sensitivity but also amplifying physically-induced false positives. 

In this context, integrating physical constraints during post-processing can help mitigate such errors. For example, running avalanche simulations under extreme conditions can generate a mask encompassing all potentially affected areas. Restricting inference to this physically plausible domain can effectively suppress false positives outside avalanche-prone zones, thereby improving both accuracy and interpretability.

The current patch-based design limits the model’s spatial awareness, as each tile captures only a limited portion of the terrain and therefore lacks broader contextual information such as entire avalanche paths or slope-scale terrain structures. Patch extraction is nevertheless required for two practical reasons. First, full Sentinel-1 scenes cover very large spatial extents that exceed the memory limits of current deep neural network architectures, making direct scene-level training infeasible on standard GPU hardware. Second, extracting overlapping patches substantially increases the number of training samples available from each scene, which is essential given the relatively limited number of avalanche events in the dataset. This tiling strategy is widely adopted in remote sensing segmentation tasks where satellite scenes are substantially larger than the spatial input limits of convolutional or transformer-based architectures. While overlapping tiles partially mitigate the loss of spatial context during inference, larger-context approaches or scene-level models could further improve spatial reasoning and will be investigated in future work.

Some error sources remain difficult to mitigate directly within the current framework. In particular, widespread snow refreezing can still produce extensive false positives, although such cases are often recognizable due to their spatial uniformity.

Overall, the proposed framework achieved strong performance, with Gaussian blending providing the best agreement with the reference masks and F2-based thresholding improving recall in more operational, recall-oriented settings.

\section{Conclusion}
\label{sec:conclusion}

This work presents an end-to-end framework for large-scale avalanche segmentation from bi-temporal Sentinel-1 SAR imagery. The main result is that a unimodal change-detection formulation based only on pre- and post-event SAR images is sufficient to achieve strong performance and, in the experiments, more consistent than multimodal variants that incorporate terrain auxiliaries.

Among the evaluated settings, $128 \times 128$ patches combined with Gaussian blending produced the best overall agreement with the reference masks, whereas F2-based threshold tuning with max blending improved recall and increased the detection of smaller avalanches. These results show that threshold selection and tile-fusion strategy have a direct effect on the balance between conservative mapping and broader event coverage.

More generally, the findings suggest that simplifying the input pipeline does not necessarily reduce performance, which is encouraging for scalable SAR-based avalanche monitoring. Future work should investigate stronger physical constraints, additional environmental context, uncertainty quantification, and broader validation across regions and acquisition conditions. The release of the manually validated multi-region dataset provides a reproducible benchmark for further research and operational development.


\section*{CRediT Authorship Contribution Statement}

\textbf{Mattia Gatti}: Conceptualization, Methodology, Supervision, Software, Investigation, Data curation, Visualization, Writing -- original draft.
\textbf{Alberto Mariani}: Data curation, Validation, Formal analysis, Visualization, Writing -- review \& editing.
\textbf{Ignazio Gallo}: Conceptualization, Methodology, Visualization, Writing -- review \& editing.  
\textbf{Fabiano Monti}: Resources, Data curation, Writing -- review \& editing.

\section*{Declaration of Competing Interest}
The authors declare that they have no known competing financial interests or personal relationships that could have appeared to influence the work reported in this paper.

\section*{Code Availability}
\label{sec:code_availability}

The code developed for this study is publicly available at \url{https://github.com/mattiagatti/avalanche-deep-change-detection}. 
The implementation is written in Python and makes use of PyTorch for deep learning and GDAL for geospatial data processing.

\section*{Data Availability}
\label{sec:data_availability}

The avalanche dataset analyzed in this study is publicly available at
\href{https://doi.org/10.5281/zenodo.15863589}{https://doi.org/10.5281/zenodo.15863589}.

\bibliographystyle{elsarticle-harv}
\bibliography{bibliography}


\end{document}